\documentclass[runningheads]{llncs}

\usepackage{eccv}

\usepackage{eccvabbrv}

\usepackage{graphicx}
\usepackage{booktabs}

\usepackage[accsupp]{axessibility}  

\usepackage[hidelinks]{hyperref}

\usepackage{orcidlink}

\usepackage{colortbl}
\usepackage{pifont}
\usepackage{makecell}
\usepackage{array}
\usepackage{tabularx}
\usepackage{bm}
\usepackage{enumitem}
\setlist[itemize]{leftmargin=*}

\usepackage{amsmath,amsfonts,bm}
\usepackage{xcolor}

\def\eqref#1{equation~\ref{#1}}

\def\1{\bm{1}}

\DeclareMathAlphabet{\mathsfit}{\encodingdefault}{\sfdefault}{m}{sl}
\SetMathAlphabet{\mathsfit}{bold}{\encodingdefault}{\sfdefault}{bx}{n}

\definecolor{okgreen}{RGB}{30,132,73}    
\definecolor{nored}{RGB}{192,57,43}      
\definecolor{neutralgray}{RGB}{127,140,141} 
\definecolor{ourshl}{RGB}{235,240,245}   
\definecolor{partialorange}{RGB}{213,94,0} 

\newcommand{\cmark}{\textcolor{okgreen}{\ding{51}}}
\newcommand{\xmark}{\textcolor{nored}{\ding{55}}}
\newcommand{\pmark}{\textcolor{partialorange}{\ding{109}}}

\newcolumntype{C}{>{\centering\arraybackslash}X}

\newcommand\blfootnote[1]{\begingroup\renewcommand\thefootnote{}\footnote{#1}\addtocounter{footnote}{-1}\endgroup}

\begin{document}

\title{SoftVTBench: A Safety-Aware Visuo-Tactile Benchmark for Physically Constrained Robotic Manipulation of Deformable Objects~\mbox{(Early Version)}}

\titlerunning{SoftVTBench: A Safety-Aware Visuo-Tactile Benchmark}

\author{
Bowen Jing\inst{1}$^{*}$ \and
Mingxin Wang\inst{1,2}$^{*}$ \and
Ruiyang Hao\inst{3} \and
Chenchen Ge\inst{1,4} \and
Hanwen Shen\inst{5} \and
Junjie He\inst{6} \and
Yang Cui\inst{7} \and
Yiming Hou\inst{1,4} \and
Weitao Zhou\inst{2,8}$^{\dagger}$ \and
Jiawei Wang\inst{8} \and
Minglei Li\inst{8} \and
Dandan Zhang\inst{9} \and
Ding Zhao\inst{10} \and
Houde Liu\inst{2} \and
Xiaofan Li\inst{11} \and
Si Liu\inst{12} \and
Ping Luo\inst{13} \and
Haibao Yu\inst{1,13}$^{\dagger}$
}

\authorrunning{B.~Jing et al.}

\institute{
$^{1}$Tuojing Intelligence,
$^{2}$Tsinghua University,
$^{3}$King's College London,
$^{4}$Southeast University,
$^{5}$Stevens Institute of Technology,
$^{6}$The Hong Kong University of Science and Technology (Guangzhou),
$^{7}$University of Manchester,
$^{8}$Simple AI,
$^{9}$Imperial College London,
$^{10}$Carnegie Mellon University,
$^{11}$Zhejiang University,
$^{12}$Beihang University,
$^{13}$The University of Hong Kong
}

\maketitle

\blfootnote{\footnotesize $^{*}$~Equal contribution.\quad
$^{\dagger}$~Corresponding author.\quad
Code and project page: \href{https://softvtbench.github.io}{softvtbench.github.io}}

\begin{abstract}
Deformable object manipulation poses challenges beyond task completion:
execution must also maintain safe physical interaction, holding the object
stably without slip or drop while avoiding excessive deformation. Existing
manipulation benchmarks, however, are predominantly success-oriented and
rarely evaluate whether a policy stays physically safe throughout execution.
We present SoftVTBench, a safety-aware visuo-tactile benchmark for physically
constrained deformable object manipulation. Built in Isaac Sim with finite
element method (FEM) deformable objects, it provides multi-view RGB
observations, RGB tactile sensing with marker motion, proprioception, and
language instructions, and defines four matched task suites over object type
(deformable vs.\ rigid) and variation axis (object vs.\ spatial). It reports
\emph{Goal Success} and \emph{Safety Success} separately; the latter also
requires no drop and peak deformation below a calibrated object-specific
threshold, measured from policy-hidden privileged FEM states. We implement
$\pi_{0.5}$-based baselines under this protocol. Experiments show that
success-only evaluation substantially overstates policy performance: a large
fraction of goal-completing rollouts violate physical safety. Tactile sensing
improves Safety Success (e.g., $21.4\%$ to $35.6\%$ on object-centric
deformable tasks) and reduces deformation during execution, while keeping
Goal Success comparable. SoftVTBench thus offers a reproducible testbed for
visuo-tactile deformable manipulation under physical interaction constraints.

\keywords{Deformable Object Manipulation \and Visuo-Tactile Sensing \and Robot Manipulation Benchmark \and Safety-Aware Evaluation}
\end{abstract}

\section{Introduction}
\label{sec:intro}

Recent progress in large-scale robot learning and foundation-model-based
policies has significantly improved the generality of robotic manipulation
across tasks, objects, and environments~\cite{walke2023bridgedata,
kim2024openvla, black2024pi0, bjorck2025groot, jang2025dreamgen,
ji2025robobrain}. However, most existing approaches are primarily evaluated
in rigid-object settings~\cite{liu2023libero, gu2023maniskill2,
nasiriany2024robocasa, mu2025robotwin}, where performance is measured by
goal achievement and the physical interaction process is largely abstracted
away. Although deformable object manipulation has recently received increasing attention~\cite{zhao2025learning, moletta2026preference, moghani2026softmimicgen}, its evaluation protocols remain largely success-oriented~\cite{huang2021plasticinelab, zhang2025modesuite}. Unlike rigid-object manipulation, deformable object manipulation inherently involves
contact-rich dynamics and material-dependent
constraints~\cite{sun2025soft}, in which successful
execution requires not only accomplishing the task but also maintaining
physically appropriate interactions, such as holding the object stably
without slip or drop and avoiding excessive deformation or damage caused by
improper grasping forces. 
Therefore, evaluating deformable object manipulation requires going beyond success-oriented metrics to explicitly assess interaction-level physical constraints throughout execution.


Evaluating such physical constraints requires robots to perceive
fine-grained contact dynamics during manipulation. While vision provides
global information about object geometry and pose, it fails to capture
critical interaction states at the gripper--object interface, including
contact pressure distribution, incipient slip, and local
deformation~\cite{huang20243d, zheng2026omnivta}. These signals are
essential for maintaining grasp stability and physical safety during
execution. Hence, vision-based policies often struggle to regulate
grasping forces reliably in contact-rich scenarios, especially under varying
object compliance and frictional uncertainty~\cite{he2025foar}. Tactile
sensing complements vision by providing direct feedback on local physical
interactions, enabling more accurate perception of contact dynamics and more
reliable regulation of manipulation behavior~\cite{suresh2024neuralfeels, fan2024vitactip, liu2026learning}. Consequently, a benchmark for physically constrained deformable object manipulation should support visuo-tactile observations, enabling evaluation beyond vision-only perception.

\begin{figure}[t]
    \centering
    \includegraphics[width=1.0\linewidth]{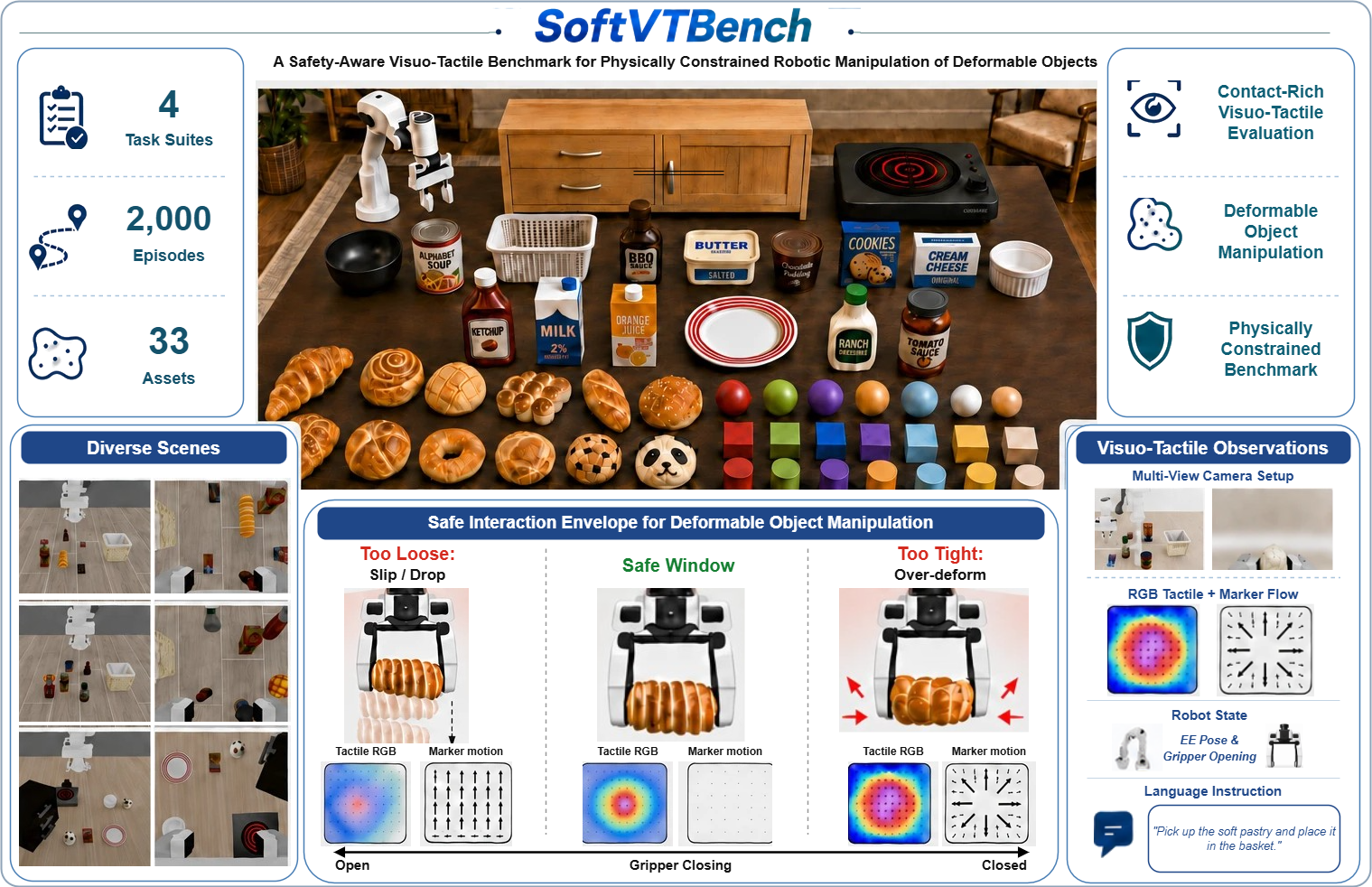}
    \caption{
    \textbf{Overview of SoftVTBench.}
    SoftVTBench is a safety-aware visuo-tactile benchmark for physically constrained deformable-object manipulation, comprising four task suites, $2{,}000$ episodes, 33 assets, and diverse tabletop scenes.
    Each episode provides multi-view RGB observations, dual-finger tactile observations with RGB images and marker-flow signals, robot proprioception, and language instructions.
    The safe-interaction envelope illustrates the core premise of SoftVTBench: unlike success-only benchmarks that only check whether a task is completed, SoftVTBench evaluates whether completion is achieved under physically safe contact.
    A grasp that is too loose may cause slip or drop, whereas an overly tight grasp may complete the placement task but over-deform the object; safe manipulation therefore lies in the intermediate regime that preserves grasp stability while limiting deformation.
    }
    
    \label{fig:overview}
\end{figure}

Despite rapid progress in robotic manipulation, tactile sensing, and deformable object manipulation, existing benchmarks each cover only part of what physically constrained deformable manipulation requires. Manipulation benchmarks such as LIBERO~\cite{liu2023libero}, RoboTwin~\cite{mu2025robotwin}, RoboCasa~\cite{nasiriany2024robocasa}, and SIMPLER~\cite{gu2024simpler} standardize evaluation for general policies, but target rigid-object tasks with success-oriented metrics. Tactile and visuo-tactile benchmarks such as UniVTAC~\cite{chen2026univtac}, TacO~\cite{zorin2026taco}, and Tabero~\cite{wu2026tabero} investigate tactile perception and contact-rich manipulation, but not deformation-aware evaluation. Deformable object manipulation benchmarks~\cite{lin2020softgym, zhang2025modesuite, greenland2024sograb} evaluate manipulation in deformable environments or assess grasp-induced deformation, rather than policy behavior under deformation-bounded interaction constraints. The intersection of these three directions therefore remains underexplored.

To address this gap, we present SoftVTBench, a safety-aware visuo-tactile benchmark for physically constrained
deformable object manipulation. Figure~\ref{fig:overview} provides an overview of SoftVTBench, including its benchmark scale, diverse scenes and assets, multimodal visuo-tactile observation interface, and safe interaction envelope for deformable-object manipulation. 
SoftVTBench provides a unified closed-loop evaluation environment with realistic deformable assets, multi-view visual observations, dual-finger RGB tactile sensing, proprioception, and language instructions, together with nearly $2{,}000$ collected episodes for visuo-tactile deformable manipulation. Its object-centric and spatial tasks are paired with matched rigid-control suites, separating basic manipulation competence from safe deformable-object interaction.
Unlike success-only evaluation, SoftVTBench evaluates physical safety using policy-hidden privileged Finite Element Method (FEM)~\cite{zienkiewicz1977finite} states to detect drop/slip events and identify excessive deformation beyond calibrated object-specific thresholds.
Based on this benchmark, we
implement $\pi_{0.5}$-based baseline policies~\cite{physical2025pi} and
compare observation modalities under a shared physical and task setup.
Experiments show that success-only evaluation substantially overstates policy
performance, as a large fraction of goal-completing rollouts violate physical
safety, and that adding tactile sensing improves Safety Success and reduces
object deformation during execution while keeping Goal Success comparable.

In summary, this work makes three contributions:
\begin{itemize}
\item We formulate safety-aware evaluation for deformable object
manipulation, in which a policy must complete the task while keeping the
object stably grasped and its deformation below a calibrated, object-specific
threshold, exposing unsafe completions that are hidden by success-only
evaluation.
\item We introduce a visuo-tactile Isaac Sim benchmark with FEM-simulated
deformable assets, multi-view visual and RGB tactile observations,
object-centric and spatial task suites paired with matched rigid-control
suites, and privileged-state safety evaluation.
\item We implement $\pi_{0.5}$-based baselines and show that success-only
metrics substantially overstate policy performance, and that tactile sensing
improves Safety Success while reducing excessive deformation.
\end{itemize}
\section{Related Work}
\label{sec:related-work}

\paragraph{Robotic Manipulation Benchmarks}
Manipulation benchmarks have established standard protocols for measuring progress in robot learning. Meta-World~\cite{yu2020metaworld} and RLBench~\cite{james2020rlbench} introduced multi-task and language-conditioned suites, LIBERO~\cite{liu2023libero} evaluates lifelong knowledge transfer across 130 language-conditioned tasks, and CALVIN~\cite{mees2021calvin} focuses on long-horizon language-conditioned control. Later benchmarks scale task, scene, and embodiment diversity: RoboCasa~\cite{nasiriany2024robocasa} procedurally generates household manipulation, RoboTwin~\cite{mu2025robotwin} and RoboTwin~2.0~\cite{chen2025robotwin2} target bimanual manipulation with synthetic data generation and domain randomization, THE COLOSSEUM~\cite{pumacay2024colosseum} stress-tests robustness under visual and physical perturbations, and SIMPLER~\cite{gu2024simpler} provides simulation-based evaluation whose policy rankings predict real-world performance. Their success predicates, however, are primarily terminal and kinematic: whether an object reaches a target state or a subtask is completed. Safety-oriented benchmarks have begun to evaluate safety alongside task success~\cite{ji2023safetygym, zhang2025safevla, fan2026safevlabench, huang2026safemanip}, but mainly target constraint violations, semantic hazards, or temporal-logic properties for rigid-object or general policies. SoftVTBench instead focuses on process-level physical safety induced by contact with deformable objects, where task completion alone does not characterize interaction quality during execution. Table~\ref{tab:positioning} summarizes its positioning relative to representative manipulation, deformable-object, and visuo-tactile benchmarks.

\begin{table*}[t]
\centering
\caption{
Comparison with representative manipulation, deformable-object, and
visuo-tactile benchmarks.
\cmark: direct support; \pmark: partial support; \xmark: not supported.
\emph{Tactile Observation}: tactile signals in the policy observation;
\emph{Deformable Objects}: 3D deformable target objects;
\emph{Deformation GT}: deformation measured from physical simulation ground
truth; \emph{Safety Metric}: evaluation beyond goal completion;
\emph{Rigid--Soft Controls}: matched rigid/deformable task suites.
}
\label{tab:positioning}
\renewcommand{\arraystretch}{1.10}
\setlength{\tabcolsep}{3pt}
\footnotesize
\begin{tabularx}{\textwidth}{@{}l*{5}{>{\centering\arraybackslash}X}@{}}
\toprule
\textbf{Benchmark}
& \scriptsize\textbf{Tactile Observation}
& \scriptsize\textbf{Deformable Objects}
& \scriptsize\textbf{Deformation GT}
& \scriptsize\textbf{Safety Metric}
& \scriptsize\textbf{Rigid--Soft Controls} \\
\midrule
LIBERO~\cite{liu2023libero}
& \xmark & \xmark & \xmark & \xmark & \xmark \\
SoftGym~\cite{lin2020softgym}
& \xmark & \pmark & \pmark & \xmark & \xmark \\
MoDeSuite~\cite{zhang2025modesuite}
& \xmark & \cmark & \pmark & \xmark & \xmark \\
SoGraB~\cite{greenland2024sograb}
& \xmark & \cmark & \cmark & \pmark & \xmark \\
UniVTAC~\cite{chen2026univtac}
& \cmark & \pmark & \xmark & \xmark & \xmark \\
ManiFeel~\cite{luu2025manifeel}
& \cmark & \xmark & \xmark & \xmark & \xmark \\
Tabero~\cite{wu2026tabero}
& \cmark & \pmark & \xmark & \pmark & \xmark \\
\midrule
\rowcolor{ourshl}
\textbf{\textsc{SoftVTBench}}
& \cmark & \cmark & \cmark & \cmark & \cmark \\
\bottomrule
\end{tabularx}
\normalsize
\end{table*}

\paragraph{Deformable-Object Manipulation}
Deformable-object manipulation has been studied across simulation environments, task suites, and learning methods, where object states are high-dimensional and governed by complex dynamics~\cite{zhu2022challenges}. SoftGym~\cite{lin2020softgym} provides reinforcement-learning tasks for cloth, ropes, and other deformable objects, while PlasticineLab~\cite{huang2021plasticinelab} and DaXBench~\cite{chen2022daxbench} use differentiable soft-body simulation to support learning and optimization. DEDO~\cite{antonova2021dedo} and GarmentLab~\cite{lu2024garmentlab} extend this line with dynamic cloth and garment tasks, ManiSkill2~\cite{gu2023maniskill2} includes soft-body tasks in a broader manipulation suite, MoDeSuite~\cite{zhang2025modesuite} targets mobile manipulation with deformable objects, and recent foundation-model work such as DeMaVLA~\cite{su2026demavla} reflects growing interest in scalable policies for deformable manipulation. Many of these settings focus on thin-shell or low-dimensional deformables such as cloth, rope, and plasticine, where deformation is often part of the task state or objective.
SoftVTBench addresses a complementary regime in which deformation is not the goal, but a physical quantity that must remain bounded while another manipulation objective is completed. This distinction is important for everyday manipulation of volumetric soft objects such as food, packaging, and soft containers, where the robot should grasp and transport the object without changing its physical condition more than necessary~\cite{zhu2022challenges}. The closest prior work measures deformation from simulation ground truth: SoGraB~\cite{greenland2024sograb} scores gripper-induced deformation during soft grasping, while DefGraspSim~\cite{huang2022defgraspsim} and DefGraspNets~\cite{huang2023defgraspnets} evaluate FEM stress and deformation of candidate grasps on 3D deformable objects. These methods provide valuable deformation-aware grasp evaluation, whereas SoftVTBench evaluates policy behavior under deformation-bounded interaction constraints throughout a full manipulation task with visuo-tactile observations.

\paragraph{Visuo-Tactile Sensing and Policy Learning}

Tactile sensing provides local physical information that vision alone cannot observe, including contact geometry, shear, slip, and compression~\cite{yuan2017gelsight, si2022taxim, suresh2024neuralfeels}. GelSight~\cite{yuan2017gelsight} introduced high-resolution optical tactile sensing, and simulation tools such as Taxim~\cite{si2022taxim}, FOTS~\cite{zhao2024fots}, TacEx~\cite{nguyen2024tacex}, TacSL~\cite{akinola2025tacsl}, and DiffTactile~\cite{si2024difftactile} make tactile observations more accessible for learning-based manipulation. Recent policy and benchmark work increasingly treats tactile feedback as a first-class observation stream: VTLA~\cite{zhang2025vtla}, OmniVTLA~\cite{cheng2025omnivtla}, VLA-Touch~\cite{bi2025vlatouch}, and Tactile-VLA~\cite{huang2025tactilevla} incorporate tactile signals into vision-language-action policies; ManiFeel~\cite{luu2025manifeel} benchmarks visuo-tactile manipulation policy learning; UniVTAC~\cite{chen2026univtac} provides a unified simulation platform for visuo-tactile data generation, representation learning, and benchmarking across contact-rich tasks; TacO~\cite{zorin2026taco} compares tactile sensor modalities under a task-driven imitation-learning protocol; and AT-VLA~\cite{li2026atvla} studies adaptive tactile injection and fast tactile reaction in VLA models. These works show the value of tactile feedback for contact-rich manipulation. However, they primarily study rigid-object or general contact-rich settings, leaving open whether tactile feedback helps policies satisfy deformation-bounded safety constraints when the true object deformation state is hidden. SoftVTBench targets this question directly by evaluating whether visual and tactile feedback enable policies to avoid unsafe over-compression, slip, and drop while still completing manipulation goals.

\section{SoftVTBench}
\label{sec:benchmark}





\subsection{Benchmark Design}
\label{sec:benchmark_design}

SoftVTBench is a safety-aware visuo-tactile benchmark for physically
constrained deformable object manipulation. It evaluates closed-loop robot
policies under two coupled requirements: completing the manipulation goal and
maintaining safe physical interaction throughout execution. A rollout is
considered physically safe only if the object remains stably grasped without
slip or drop, and its peak deformation stays below a calibrated,
object-specific threshold. By reporting \emph{Goal Success} and
\emph{Safety Success} separately, SoftVTBench exposes goal-complete but
physically unsafe rollouts that success-only evaluation hides.

\begin{figure}[t]
    \centering
    \includegraphics[width=1\linewidth]{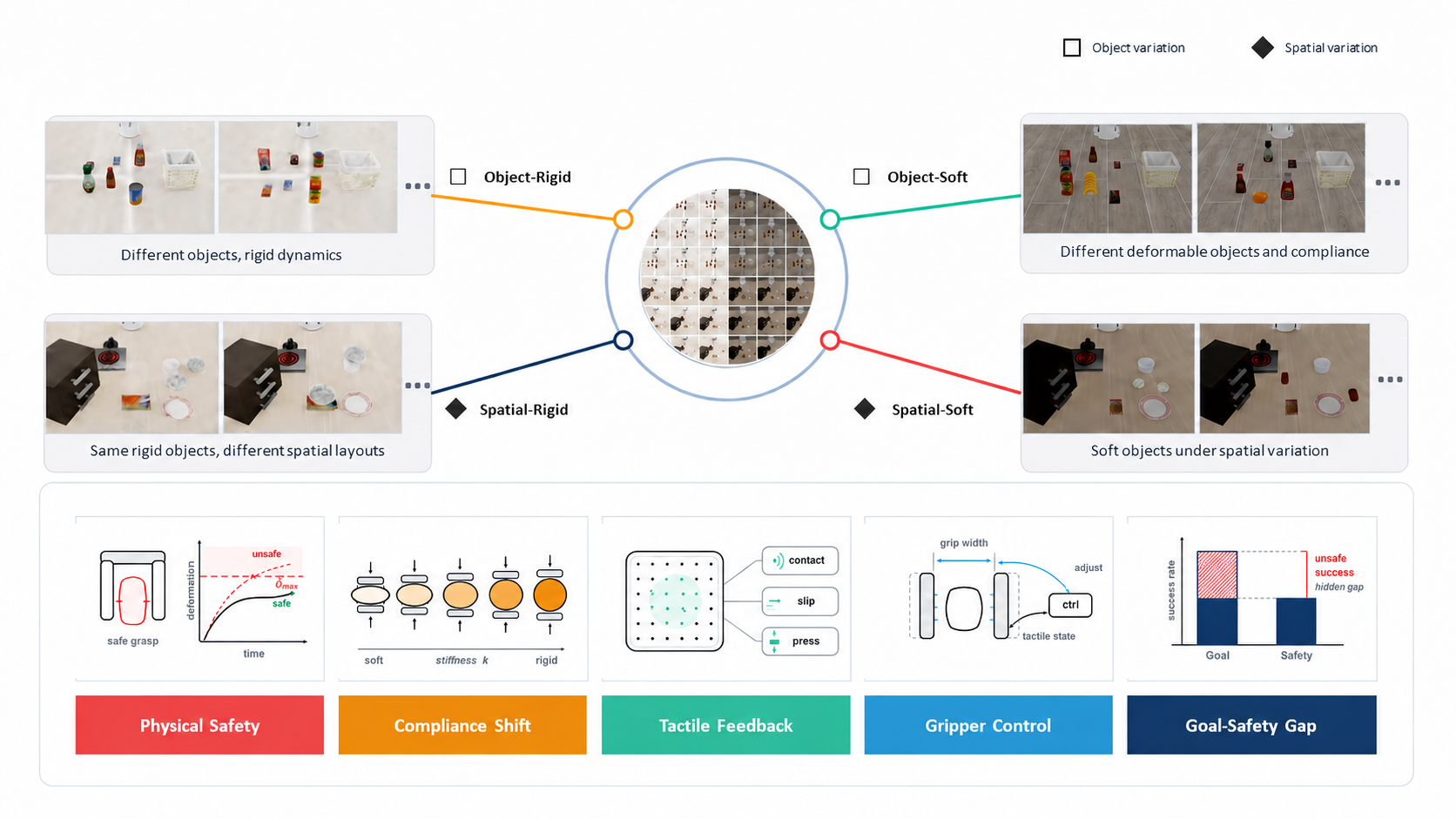}
\caption{
SoftVTBench organizes evaluation into four matched task suites defined by object type and variation type: \textsc{Object-Soft}, \textsc{Spatial-Soft}, \textsc{Object-Rigid}, and \textsc{Spatial-Rigid}.
The benchmark studies grasp-and-place manipulation under a hidden safe interaction envelope, where the robot must maintain stable contact without slip or drop while keeping peak FEM deformation below a calibrated object-specific threshold.
The bottom panels summarize the key factors: physical deformation safety, compliance shift, tactile feedback, gripper control, and the \emph{Goal--Safety Gap}, where goal-completing rollouts remain physically unsafe under success-only evaluation.
}
    \label{fig:softvtbench_overview}
\end{figure}

SoftVTBench instantiates this evaluation protocol in Isaac Sim with simulated deformable objects based
on finite element method soft-body dynamics and a Franka Panda arm with a parallel-jaw gripper
carrying GelSight Mini tactile sensors on both fingers, as illustrated in Figure~\ref{fig:softvtbench_overview}. The benchmark provides synchronized
third-person and wrist RGB observations, tactile RGB images, marker-motion
fields, proprioception, and a standardized end-effector and gripper action
interface. It contains object-centric and spatial deformable manipulation
suites, together with matched rigid-control suites that separate
deformable-object safety from basic manipulation competence and robustness to
spatial variation.

At each control step, the policy receives only its designated observations and
outputs an end-effector command and a gripper command. The simulator advances
the robot, object, and contact states, producing the next observations in a
closed-loop interaction process. In parallel, SoftVTBench records privileged
physical states, including FEM nodal positions, object poses, contact status,
and drop events. These states are used only by the evaluator, requiring
policies to infer contact conditions, incipient slip, and material compliance
from observable visual, tactile, and proprioceptive inputs.

\subsection{Observation and Action Space}
\label{sec:obs_action}

SoftVTBench provides a standardized multimodal observation-action interface
for closed-loop policy evaluation. At each control step, the policy receives
visual, tactile, and proprioceptive observations together with a per-task
language instruction, and outputs an end-effector command and a gripper
command.

The visual observation consists of two RGB views: a third-person camera that
captures the tabletop layout, target region, manipulated object, and
distractors, and a wrist-mounted camera that provides close-range views
during approach, grasping, and transport. The tactile observation is obtained
from simulated GelSight Mini sensors~\cite{yuan2017gelsight} mounted on both
gripper fingers. Each tactile sensor provides an RGB tactile image and a
marker-motion field, which capture local contact geometry, elastomer
deformation, and shear motion at the gripper--object interface. These tactile
signals provide interaction-state information that is often occluded or
ambiguous in external visual observations.

The proprioceptive state consists of the end-effector pose, arm joint states,
and current gripper width. Each task is additionally specified by a
natural-language instruction identifying the target object and goal region.
All observation streams are synchronized at 20\,Hz. Formally, the
policy-visible observation can be written as
\begin{equation}
o_t =
\{I_t^{\mathrm{third}}, I_t^{\mathrm{wrist}},
\mathcal{T}_t, p_t, \ell\},
\end{equation}
where $I$ denotes the third-person and wrist RGB images, $\mathcal{T}_t$ the
tactile observation from both fingers---comprising the tactile RGB images and
marker-motion fields of the left and right GelSight sensors---$p_t$ the
proprioceptive state, and $\ell$ the task instruction.

The action space contains an absolute end-effector pose target and a scalar
gripper command:
\begin{equation}
a_t = (\mathbf{x}_t^{ee}, \boldsymbol{\theta}_t^{ee}, g_t),
\end{equation}
where $\mathbf{x}_t^{ee}$ is the 3D end-effector position,
$\boldsymbol{\theta}_t^{ee}$ is the axis-angle orientation, and $g_t$ denotes
the gripper command. SoftVTBench supports both binary open/close commands and
continuous closure targets, with aligned trajectory encodings for controlled
ablation of gripper-action granularity. Detailed sensor resolutions,
rendering pipelines, synchronization rates, and action parameterization are
provided in Appendix~\ref{app:implementation}.

\subsection{Task Suites}
\label{sec:tasks}

SoftVTBench defines four task suites organized along two axes: object type and task variation, as summarized in Table~\ref{tab:task_suites}. The object-type axis contrasts deformable objects with
matched rigid-control objects, while the task-variation axis contrasts
object-centric variation with spatial variation. This design lets us separate
deformable-object safety from basic manipulation competence and robustness to
spatial variation.

The deformable suites use FEM-simulated objects,
including bakery-style objects and geometric primitives. These objects vary
in shape, size, and appearance, producing object-specific contact responses,
manipulation difficulty, and safety thresholds. Because the safe interaction
range is not directly observable from vision alone, policies must infer safe
manipulation behavior from visual, tactile, and proprioceptive feedback.
Detailed asset statistics and material parameters are provided in
Appendix~\ref{app:calibration}.

\begin{table}[t]
\centering
\caption{Task suites in SoftVTBench. The four suites span deformable and rigid objects under object-level and spatial variation, separating manipulation performance from physical safety across different sources of task diversity.}
\small
\begin{tabular}{llll}
\toprule
\textbf{Suite} & \textbf{Object Type} & \textbf{Variation} & \textbf{Purpose} \\
\midrule
\textsc{Object-Soft} & Deformable & Object & Safe deformable manipulation \\
\textsc{Spatial-Soft} & Deformable & Spatial & Safe manipulation under layout variation \\
\textsc{Object-Rigid} & Rigid & Object & Basic manipulation control \\
\textsc{Spatial-Rigid} & Rigid & Spatial & Spatial-variation control \\
\bottomrule
\end{tabular}
\label{tab:task_suites}
\end{table}

In \textsc{Object-Soft}, the robot must grasp a deformable object and place
it into a target container without dropping or excessively deforming it. The
scene layout and target region are fixed, while the manipulated object varies
across tasks. This suite evaluates whether a policy can adapt its interaction
to object-level differences in geometry and contact response.

In \textsc{Spatial-Soft}, the robot performs the same grasp-and-place
objective under spatial variation. Each scene contains two visually identical
instances of the same deformable object, and the language instruction
specifies which instance to manipulate; tasks come in mirrored pairs that
share the same physical layout. The policy must therefore ground the
instruction to the correct instance and complete the transfer safely under
changing object and target placements.

The two rigid-control suites, \textsc{Object-Rigid} and \textsc{Spatial-Rigid},
follow the same object-centric and spatial task structures using rigid
LIBERO-style objects~\cite{liu2023libero}. These suites serve as diagnostic
controls: they test whether a policy has basic manipulation and
spatial-variation capability before the additional safety constraints
introduced by deformable objects are evaluated.

\subsection{Evaluation Protocol and Metrics}
\label{sec:eval}

SoftVTBench evaluates policies through closed-loop rollouts in the simulator.
For each task suite, all methods are evaluated on the same evaluation episodes
with fixed initial states and fixed seeds. During execution, the policy
receives only its designated observations and outputs robot actions.
Privileged simulator states, including object pose, contact status, drop
events, and FEM deformation, are recorded only for evaluation and are never
exposed to the policy.

We report two main metrics: \emph{Goal Success} and \emph{Safety Success}.
Goal Success measures whether the task objective is completed. In our
grasp-and-place tasks, a rollout is counted as goal-successful if the target
object is placed inside the target region or container and remains there for
a short terminal horizon.

Safety Success further requires the task to be completed without unsafe
physical interaction. For deformable-object suites, the evaluator computes
object deformation at each timestep from the hidden FEM state and records the
peak deformation over the full rollout:
\begin{equation}
D_{\mathrm{peak}} = \max_t D(t),
\end{equation}
where $D(t)$ is the FEM-RMS deformation after removing global rigid-body
motion, reported as a percentage of the object bounding-box diagonal.

A deformable-object rollout is counted as safety-successful if
\begin{equation}
\mathrm{Safety\ Success}
=
\mathrm{Goal\ Success}
\;\wedge\;
\mathrm{NoDrop}_{\mathrm{episode}}
\;\wedge\;
\left(D_{\mathrm{peak}} \leq \tau_o\right),
\end{equation}
where $\tau_o$ is the calibrated, object-specific safety threshold.
$\mathrm{NoDrop}_{\mathrm{episode}}$ requires the object to remain under
stable manipulation throughout the episode, and is violated by any transient
drop, workspace escape, or loss of stable containment. For the rigid-control
suites, the deformation term is inactive, and Safety Success reduces to Goal
Success under the NoDrop condition. Over $N$ evaluation episodes, each metric
is reported as its rate:
\begin{equation}
\begin{aligned}
\mathrm{Goal\ Success\ Rate} &= \frac{1}{N}\sum_{i=1}^{N}\mathrm{Goal\ Success}^{(i)},\\[-2pt]
\mathrm{Safety\ Success\ Rate} &= \frac{1}{N}\sum_{i=1}^{N}\mathrm{Safety\ Success}^{(i)}.
\end{aligned}
\end{equation}

Safety Success is therefore stricter than Goal Success: a rollout may complete
the task while still violating physical safety through transient drop,
unstable contact, or excessive deformation. The gap between the two metrics
captures unsafe task completions that success-only evaluation hides.
\section{Experiments}

We evaluate SoftVTBench on physically constrained deformable object manipulation. Our goal is to measure not only whether a policy completes the task, but whether it does so while maintaining safe physical interaction with the object. We compare policies across object-centric and spatial settings, and report both task-level and safety-aware success.

\subsection{Baselines}

We compare two $\pi_{0.5}$-based policy settings.

\textbf{$\pi_{0.5}$-Vision}~\cite{physical2025pi}.
The vision-only policy takes as input a third-person RGB image, a wrist RGB image, robot proprioceptive state, and the language instruction. The gripper command is represented as a binary open-close action. This design avoids providing the vision-only policy with continuous gripper-width bounds obtained from contact calibration, which would otherwise introduce implicit interaction information beyond visual observations.

\textbf{$\pi_{0.5}$-Visuo-Tactile.}
The visuo-tactile policy uses the same visual and proprioceptive inputs as $\pi_{0.5}$-Vision, and additionally receives tactile observations from both gripper fingers. For tactile RGB observations, we use an 8-frame history from each finger and concatenate the left and right tactile histories into a single $4 \times 4$ grid image. This tactile image is encoded by the same visual encoder used for RGB observations. We also include tactile marker motion as a low-dimensional contact signal: marker motion from both fingers is concatenated and provided as a history sequence, allowing the policy to observe recent local contact deformation and shear. The gripper command is represented as a continuous gripper-width action.

Both policies output a 7D action consisting of a 3D end-effector position, a 3D orientation, and a 1D gripper command. We train all policies by LoRA fine-tuning $\pi_{0.5}$ with an action horizon of 50, using 8 NVIDIA A100 GPUs, a global batch size of 256, and 7k training steps by default. During evaluation, the policy predicts an action chunk and executes 10 steps before replanning.






\subsection{Main Results}


\begin{table*}[t]
    \centering
    \caption{Main results on SoftVTBench. VO is the vision-only $\pi_{0.5}$ policy and VT the visuo-tactile one. Goal Success measures task completion; Safety Success additionally requires the rollout to satisfy physical safety constraints.}
    \begin{tabular}{llcc}
        \toprule
        Suite & Method & Goal Success & Safety Success \\
        \midrule
        Object-Rigid & VO & 38.8\% & -- \\
        Object-Rigid & VT & 32.4\% & -- \\
        Spatial-Rigid & VO & 56.4\% & -- \\
        Spatial-Rigid & VT & 63.4\% & -- \\
        Object-Soft & VO & 70.4\% & 21.4\% \\
        Object-Soft & VT & 71.8\% & 35.6\% \\
        Spatial-Soft & VO & 74.2\% & 32.6\% \\
        Spatial-Soft & VT & 84.2\% & 44.6\% \\
        \bottomrule
    \end{tabular}
    \label{tab:main_results}
\end{table*}

From Table~\ref{tab:main_results}, tactile information does not consistently improve performance on rigid-object tasks. On Object-Rigid, VT reaches 32.4\% Goal Success against 38.8\% for VO, whereas on Spatial-Rigid it reaches 63.4\% against 56.4\%. Vision and proprioception therefore already supply most of the information these tasks require, and in the absence of deformation-related safety constraints the additional tactile RGB and marker-motion streams bring little benefit while introducing multimodal noise. They do not show that tactile sensing is ineffective, only that its value lies elsewhere than in rigid-object goal success.

In contrast, VT shows clearer advantages on deformable-object tasks. On Object-Soft, VO and VT achieve similar Goal Success, 70.4\% and 71.8\%, while VT improves Safety Success from 21.4\% to 35.6\%. On Spatial-Soft, VT improves both metrics, with Safety Success increasing from 32.6\% to 44.6\%. Tactile sensing thus does not merely help the object reach the target region; it improves contact regulation during execution. By using tactile RGB and marker motion, VT reduces unsafe goal-completing behaviors such as slippage, over-compression, and unstable grasping. For deformable-object manipulation, Safety Success is therefore a more faithful measure of policy quality than Goal Success alone.

\begin{figure*}[!t]
    \centering
    \includegraphics[width=\linewidth]{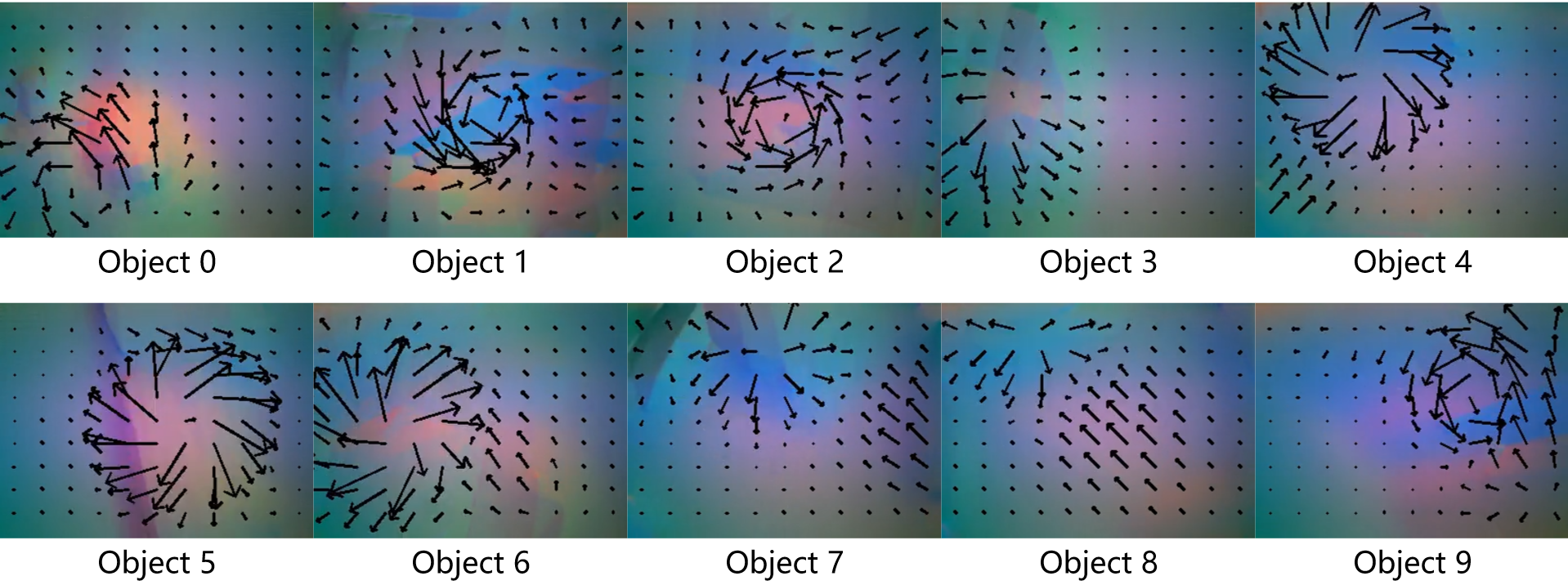}
    \caption{
    Representative tactile RGB observations from the ten Object-Soft assets in SoftVTBench. Each panel shows a tactile observation with marker-motion overlay during grasping. Different deformable objects induce diverse local contact and shear patterns, reflecting their distinct geometry, compliance, and contact response.
    }
    \label{fig:tactile_rgb_examples}
\end{figure*}

Figure~\ref{fig:tactile_rgb_examples} shows qualitatively why tactile observations are informative for deformable-object manipulation. Different Object-Soft assets produce distinct tactile RGB and marker-motion patterns during grasping, reflecting object-specific geometry, compliance, and local contact response. These interaction cues are difficult to infer from external RGB observations alone, but are captured directly by tactile sensing at the gripper--object interface.

\subsection{Deformation Analysis}


\begin{table*}[t]
    \centering
    \caption{
    FEM-RMS deformation distribution over all deformable-object rollouts in
    SoftVTBench, reported as percentages of the object bounding-box diagonal.
    We report the mean, the 5th and 95th percentiles (P5, P95), and the
    median; lower values indicate safer physical interaction. Across both
    suites, VT shifts the entire distribution downward relative to VO---%
    including the P95 tail that corresponds to severe over-compression---%
    indicating fewer and milder unsafe contacts rather than a change confined
    to the mean.
    }
    \renewcommand{\arraystretch}{1.12}
    \setlength{\tabcolsep}{6pt}
    \begin{tabular}{llcccc}
        \toprule
        Suite & Method & Mean & P5 & Median & P95 \\
        \midrule
        Object-Soft  & VO & 16.10\% & 4.30\% & 10.65\% & 44.70\% \\
        Object-Soft  & VT & 15.12\% & 3.90\% & 8.67\%  & 38.81\% \\
        Spatial-Soft & VO & 13.16\% & 5.16\% & 10.75\% & 28.96\% \\
        Spatial-Soft & VT & 11.58\% & 4.75\% & 9.67\%  & 26.56\% \\
        \bottomrule
    \end{tabular}
    \label{tab:deformation_distribution}
\end{table*}

The deformation statistics in Table~\ref{tab:deformation_distribution} explain why VT achieves higher Safety Success on deformable-object tasks. On Object-Soft, VT reduces the mean deformation from 16.10\% to 15.12\%, the median from 10.65\% to 8.67\%, and P95 from 44.70\% to 38.81\%. On Spatial-Soft, it lowers the mean from 13.16\% to 11.58\%, the median from 10.75\% to 9.67\%, and P95 from 28.96\% to 26.56\%.

Tactile sensing therefore improves safety not only by raising the threshold-based Safety Success metric, but also by shifting the continuous deformation distribution toward safer interactions. The lower median indicates better typical rollout quality, while the lower P95 shows that tactile feedback also suppresses high-deformation tail cases such as severe compression or unstable contact.

Read together with the main results in Table~\ref{tab:main_results}, this deformation analysis supports the central claim of SoftVTBench: tactile sensing matters most when a policy has to regulate local physical interaction with a deformable object. VT may not produce large gains in Goal Success, but it manipulates more safely by reducing deformation and avoiding unsafe yet goal-completing behaviors.

\subsection{Goal Success vs. Safety Success}

\begin{figure}[t]
    \centering
    \includegraphics[width=1\linewidth]{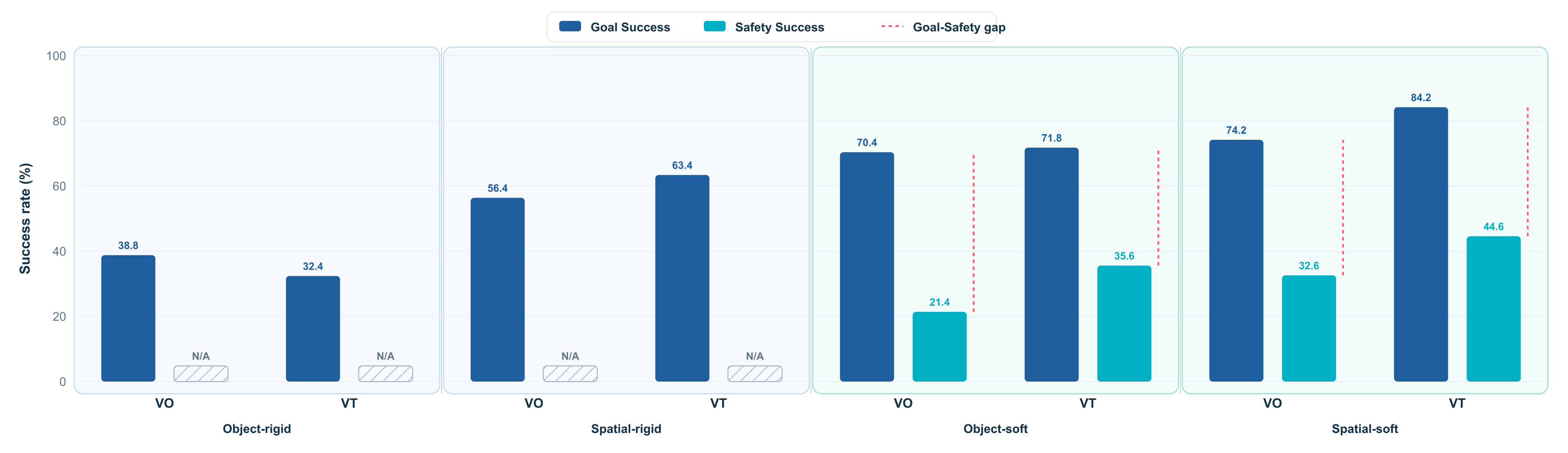}
    \caption{Goal Success and Safety Success across rigid and deformable task suites. Goal Success measures task completion, while Safety Success further requires satisfying physical safety constraints. For rigid suites, deformation-based safety constraints do not apply and are marked N/A. For soft suites, the gap between the two metrics indicates unsafe goal completions caused by excessive deformation, dropping, or slip.}
    \label{fig:goal_vs_safety}
\end{figure}

Figure~\ref{fig:goal_vs_safety} compares Goal Success and Safety Success across task suites and policy settings. The gap between the two metrics captures unsafe goal completions: rollouts where the object reaches the target but the interaction violates safety constraints. It shows why success-only evaluation is insufficient for deformable object manipulation, since a policy must not only reach the target state but also maintain appropriate contact throughout the rollout.

\section{Conclusion}
\label{sec:conclu}

We presented SoftVTBench, a safety-aware visuo-tactile benchmark for deformable object manipulation. By separating Goal Success from Safety Success and measuring FEM-RMS deformation, SoftVTBench reveals unsafe goal-completing behaviors that success-only evaluation misses. Our experiments show that tactile sensing provides limited and inconsistent gains on rigid tasks, but substantially improves safety on deformable tasks by reducing unsafe contact and object deformation.

SoftVTBench still has limitations. It covers a limited set of assets and task variations, and its deformable-object simulation cannot fully capture the complexity of real-world soft-body dynamics. We also evaluate only $\pi_{0.5}$-based baselines, leaving broader policy comparisons, such as world-action models, to future work, together with more diverse deformable assets and more realistic deformation and contact modeling.

%
%
\clearpage
\bibliographystyle{splncs04}
\bibliography{main}

\newpage
\setcounter{section}{0}
\appendix




\section{Implementation Details}
\label{app:implementation}

SoftVTBench is implemented in Isaac Sim~4.5.0 with Isaac Lab~0.41.3 and the
GPU-accelerated PhysX~5 pipeline. The simulator runs physics at 60\,Hz with
a control decimation of 3, resulting in a 20\,Hz control and logging rate.
All visual, tactile, proprioceptive, action, and privileged-state streams are
synchronized at this rate. Table~\ref{tab:implementation_details} summarizes
the robot, control, sensing, simulation, and hardware configuration.

\begin{table}[h]
\centering
\captionsetup{justification=centering}
\caption{Simulation and sensing configuration.}
\renewcommand{\arraystretch}{1.12}
\setlength{\tabcolsep}{4pt}
\small
\begin{tabular}{@{}p{0.34\linewidth}p{0.60\linewidth}@{}}
\toprule
\textbf{Item} & \textbf{Value} \\
\midrule
Simulator & Isaac Sim 4.5.0 / Isaac Lab 0.41.3, PhysX 5 GPU pipeline \\
Physics / control rate & 60\,Hz physics, decimation 3, 20\,Hz control \\
Robot & Franka arm with Panda parallel-jaw gripper \\
Controller & Task-space differential inverse kinematics \\
End-effector action & Absolute pose target: 3D position and 3D axis-angle orientation \\
Gripper action & Normalized closure command; continuous and binary encodings \\
Finger friction & Static $\mu_s=1.5$, dynamic $\mu_d=1.2$, max combine mode \\
Camera views & Third-person $1024{\times}1024$ and wrist $512{\times}512$, resized to $224{\times}224$ \\
Tactile sensor & GelSight Mini via TacEx; Taxim optics and FOTS markers \\
Tactile streams & Tactile RGB and $11{\times}9$ marker-motion field, $320{\times}240$ \\
FEM model & PhysX soft body with corotational linear elasticity \\
FEM solver & Hex resolution 6, 64 position iterations, damping 2.5 \\
Collection / evaluation & $1{\times}$ NVIDIA L20 GPU per worker \\
Training & $8{\times}$ NVIDIA A100-80GB / A800 GPUs \\
\bottomrule
\end{tabular}
\normalsize
\label{tab:implementation_details}
\end{table}

\section{Task Suite Details}
\label{app:tasks}

\subsection{Task Suite Summary}

SoftVTBench contains four task suites arranged as a matched $2\times2$
design over object type and variation axis. The object type is either rigid
or deformable, and the variation axis is either object identity or spatial
layout. We refer to the four suites as Object-Rigid, Object-Soft,
Spatial-Rigid, and Spatial-Soft.

The deformable suites contain $10$ tasks with $50$ expert demonstrations per
task. The rigid-control suites are built by re-executing and filtering LIBERO
demonstrations under the same SoftVTBench sensing and recording stack. All
suites share the same robot embodiment, camera views, tactile sensing
interfaces, rollout API, and evaluation protocol.

\begin{table}[h]
\centering
\caption{Statistics of the four task suites in SoftVTBench, including object type, variation axis, number of tasks, demonstration trajectories, and validation episodes.}
\renewcommand{\arraystretch}{1.12}
\setlength{\tabcolsep}{3pt}
\footnotesize
\begin{tabular}{lccccc}
\toprule
\textbf{Suite} & \textbf{Object Type} & \textbf{Variation Axis} &
\textbf{\#Tasks} & \textbf{\#Demos} & \textbf{\#Val Ep.} \\
\midrule
\textsc{Object-Rigid}  & Rigid      & Object identity & 10 & 500 & 50 \\
\textsc{Spatial-Rigid} & Rigid      & Spatial layout  & 10 & 500 & 50 \\
\textsc{Object-Soft}   & Deformable & Object identity & 10 & 500 & 50 \\
\textsc{Spatial-Soft}  & Deformable & Spatial layout  & 10 & 500 & 50 \\
\midrule
\textbf{Total} & -- & -- & 40 & 2000 & 200 \\
\bottomrule
\end{tabular}
\normalsize
\label{tab:task_suite_details}
\end{table}

\subsection{Object-Soft Suite}

The Object-Soft suite evaluates deformable-object manipulation under object
identity variation. In contrast to the Spatial-Soft suite, where two visually
similar objects appear in the same scene and the target is specified by a
spatial-language cue, the Object-Soft suite keeps the task structure fixed and
varies the manipulated deformable object. Each task uses a different soft
object category or asset, and the policy must adapt its grasping strategy to
the object's geometry, compliance, contact surface, and visual appearance.

Each task contains $50$ expert demonstrations. The robot is instructed to pick
up the specified deformable object and place it into the target receptacle. The
suite therefore tests object-level generalization across soft bodies while
holding the high-level manipulation goal fixed.

\begin{table}[h]
\centering
\captionsetup{justification=centering}
\caption{Object-Soft task definitions.}
\setlength{\tabcolsep}{3pt}
\renewcommand{\arraystretch}{1.05}
\small
\begin{tabularx}{\textwidth}{@{}c>{\raggedright\arraybackslash}Xc@{}}
\toprule
\textbf{Task} & \textbf{Language Prompt} & \textbf{\#Demos} \\
\midrule
0 & pick up the pale cream round steamed-bun pastry and place it in the basket & 50 \\
1 & pick up the soft pastry and place it in the basket & 50 \\
2 & pick up the soft pastry and place it in the basket & 50 \\
3 & pick up the soft pastry and place it in the basket & 50 \\
4 & pick up the orange round bun pastry with sesame speckles and place it in the basket & 50 \\
5 & pick up the soft pastry and place it in the basket & 50 \\
6 & pick up the soft cube and place it in the basket & 50 \\
7 & pick up the soft cylinder and place it in the basket & 50 \\
8 & pick up the soft wedge and place it in the basket & 50 \\
9 & pick up the soft capsule and place it in the basket & 50 \\
\bottomrule
\end{tabularx}
\label{tab:object_soft_tasks}
\normalsize
\end{table}

\subsection{Spatial-Soft Suite}

The Spatial-Soft suite evaluates deformable-object manipulation under paired
spatial layouts. Each scene contains two visually identical pastry instances:
one is the target object and the other is a distractor. For each layout, we
define two tasks by swapping which instance is referred to in the language
instruction. Thus, the physical scene layout is shared within each pair, while
the target object is changed through the language prompt. This design tests
whether policies can ground language to the correct deformable object under
spatial ambiguity.

All Spatial-Soft demonstrations use the same robot, sensing stack, visual
setting, and replay-format recording pipeline. Each task contains $50$
expert demonstrations. The final replay-format dataset stores one HDF5 file
per task with $50$ demonstrations, together with four synchronized single-view
videos per demonstration: third-person RGB, wrist RGB, left tactile marker
video, and right tactile marker video. The $2\times2$ preview videos are used
only for inspection and are not included in the final replay-format dataset.

\begin{table}[h]
\centering
\caption{Spatial-Soft task definitions. Each pair shares the same physical layout and pastry type, but swaps the target instance through the language prompt.}
\setlength{\tabcolsep}{3pt}
\renewcommand{\arraystretch}{1.05}
\small
\begin{tabularx}{\textwidth}{@{}c>{\raggedright\arraybackslash}Xc@{}}
\toprule
\textbf{Task} & \textbf{Language Prompt} & \textbf{\#Demos} \\
\midrule
0 & pick up the right white spiral pastry with ridged frosting and place it on the plate & 50 \\
1 & pick up the left white spiral pastry with ridged frosting and place it on the plate & 50 \\
2 & pick up the right pale cream round steamed-bun pastry with black eyes and place it on the plate & 50 \\
3 & pick up the left pale cream round steamed-bun pastry with black eyes and place it on the plate & 50 \\
4 & pick up the left small dark brown oval pastry and place it on the plate & 50 \\
5 & pick up the right small dark brown oval pastry and place it on the plate & 50 \\
6 & pick up the left long golden braided pastry with orange stripes and place it on the plate & 50 \\
7 & pick up the right long golden braided pastry with orange stripes and place it on the plate & 50 \\
8 & pick up the right orange round bun pastry with sesame speckles and place it on the plate & 50 \\
9 & pick up the left orange round bun pastry with sesame speckles and place it on the plate & 50 \\
\bottomrule
\end{tabularx}
\label{tab:spatial_soft_tasks}
\normalsize
\end{table}

\section{Deformable Assets and Interaction Calibration}
\label{app:calibration}

\subsection{Deformable Object Simulation}

SoftVTBench uses 3D deformable assets represented as volumetric meshes and
simulated as PhysX GPU FEM soft bodies with corotational linear elasticity.
The assets include naturalistic bakery-style objects and simple geometric
primitives. Per-asset physical parameters, such as density, friction,
elasticity, and damping, are authored in the simulation assets and are not
provided to the policy. Before each episode, the object is settled on the
support surface to obtain a stable initial state. The expert policy and the
evaluation module use this settled state, rather than the nominal spawn pose,
to avoid errors caused by pre-grasp drift.

\subsection{Asset List}

Table~\ref{tab:asset_categories} lists the deformable assets and rigid assets used in SoftVTBench.
The bakery-style pastry assets are adapted from publicly available 3D assets
from the EXTWIN Synthesis asset library\footnote{\url{https://synthesis.extwin.com/}},
and are converted into simulation-ready deformable meshes for SoftVTBench.
The main experiments use the subset of assets that are stable under repeated
closed-loop manipulation and have valid calibration results.

\begin{table*}[t]
\centering
\caption{Asset statistics in SoftVTBench. The benchmark combines Tabero/LIBERO rigid scene assets with deformable assets adapted or procedurally generated for safety-aware visuo-tactile manipulation.}
\renewcommand{\arraystretch}{1.12}
\setlength{\tabcolsep}{3pt}
\small

\begin{tabularx}{\textwidth}{
@{}
>{\raggedright\arraybackslash}p{0.22\textwidth}
>{\centering\arraybackslash}p{0.06\textwidth}
>{\raggedright\arraybackslash}X
@{}}
\toprule
\textbf{Category} & \textbf{\#} & \textbf{Assets} \\
\midrule

Scene surfaces / tables
& 2
& floor, table \\

Articulated scene objects
& 2
& wooden cabinet, flat stove \\

Rigid manipulation objects
& 15
& black bowl, alphabet soup can, basket, BBQ sauce bottle, butter box, chocolate pudding cup, cookies box, cream cheese box, glazed porcelain ramekin, ketchup bottle, milk carton, orange juice carton, plate, salad dressing bottle, tomato sauce bottle \\

Deformable bakery-style objects
& 11
& round sesame bun, elongated braided roll, striped croissant-like pastry, small ridged bread roll, yellow rolled pastry, curved layered pastry, twisted bread stick, spiral pastry roll, oval loaf with ridges, long soft bread roll, compact golden bun \\

Deformable procedural objects
& 3
& sphere, cube, cylinder \\

\midrule
Total
& 33
& 19 Tabero/LIBERO scene and rigid assets + 14 deformable assets \\
\bottomrule
\end{tabularx}
\label{tab:asset_categories}
\normalsize
\end{table*}

\subsection{Interaction Calibration Protocol}

For each deformable object, we perform an offline interaction calibration to
estimate its feasible grasping range and deformation threshold. The lower
bound is obtained by scanning gripper closure commands from loose to tight
and executing a grasp--lift--hold routine. The smallest closure that
reliably lifts and holds the object without slip or drop is recorded as the
lower bound.

The upper calibration is obtained from a compression sweep. For each closure
command, we record the maximum stable FEM RMS deformation of the object. We
define the reference deformation of object $o$ as
$D^{\mathrm{ref}}_o$, the maximum stable FEM RMS deformation measured during
the sweep. The safety threshold is then defined as
\begin{equation}
\tau_o = \kappa D^{\mathrm{ref}}_o ,
\end{equation}
where we use $\kappa=0.5$ by default and additionally report
$\kappa \in \{0.3, 0.7\}$ for sensitivity analysis. The resulting threshold
is used only by the evaluator and is not exposed to the policy.

\end{document}